\documentclass{article}

\usepackage{microtype}
\usepackage{graphicx}
\usepackage{subcaption}
\usepackage{booktabs} % for professional tables

\usepackage{hyperref}

\usepackage[accepted]{icml2026}

\makeatletter
\newcommand{\@preregisteraffil}[1]{%
  \ifcsname the@affil#1\endcsname\else
    \ifcsname @icmlsymbol#1\endcsname\else
      \stepcounter{@affiliationcounter}%
      \newcounter{@affil#1}%
      \setcounter{@affil#1}{\value{@affiliationcounter}}%
    \fi
  \fi
}
\@preregisteraffil{sjtu}
\@preregisteraffil{sii}
\@preregisteraffil{ailab}
\@preregisteraffil{qc}
\@preregisteraffil{tju}
\@preregisteraffil{sust}
\makeatother

\usepackage{amsmath}
\usepackage{amssymb}
\usepackage{mathtools}
\usepackage{amsthm}

\usepackage[capitalize,noabbrev]{cleveref}

\theoremstyle{plain}
\newtheorem{theorem}{Theorem}[section]

\theoremstyle{definition}

\theoremstyle{remark}

\usepackage[textsize=tiny]{todonotes}

\icmltitlerunning{Scaling by Diversified Experience for Vision-Language-Action Models}

\begin{document}

\twocolumn[
  \icmltitle{Scaling by Diversified Experience for Vision-Language-Action Models}
  \icmlsetsymbol{equal}{*}
  \icmlsetsymbol{corr}{$\dagger$}
  \begin{icmlauthorlist}
    \icmlauthor{Leiyu Wang}{equal,sjtu,sii}
    \icmlauthor{Zhaofengnian Wang}{equal,sii,tju}
    \icmlauthor{Xueqi Li}{sii,sust}
    \icmlauthor{Luoyi Fan}{sjtu,sii}
    \icmlauthor{Cewu Lu}{sjtu,sii,qc}
    \icmlauthor{Nanyang Ye}{corr,sjtu,sii,ailab}
  \end{icmlauthorlist}
  
  \icmlaffiliation{sjtu}{Shanghai Jiao Tong University}
  \icmlaffiliation{sii}{Shanghai Innovation Institute}
  \icmlaffiliation{ailab}{Shanghai AI Laboratory}
  \icmlaffiliation{qc}{Noematrix Intelligence}
  \icmlaffiliation{tju}{Tongji University}
  \icmlaffiliation{sust}{Southern University of Science and Technology}

  \icmlcorrespondingauthor{Nanyang Ye}{ynylincoln@sjtu.edu.cn}

  \icmlkeywords{VLA, Pretrain, Open-Source, ICML}

  \vskip 0.3in
]



% \printAffiliationsAndNotice{\textsuperscript{*}Equal contribution.}  % no special notice (required even if empty)
\printAffiliationsAndNotice{\icmlEqualContribution}

\begin{abstract}
Vision-Language-Action models face significant challenges in real-world deployment due to the entanglement of high-level reasoning with low-level control, and the instability of policy optimization. In this paper, we introduce SyVLA, a robust VLA model trained with diversified experiences. We propose an Intention Decoupling algorithm to isolate control-relevant features from reasoning contexts and a similar-sample guided RL pipeline to stabilize policy updates and mitigate distribution shift. Extensive experiments on real-world robotic tasks and multi-modal benchmarks demonstrate that SyVLA achieves superior task success rates and stronger out-of-distribution generalization compared to existing methods, while effectively preserving core vision-language capabilities. Codes and Datasets is released on \href{https://sy-vla.github.io/}{project page}.
\end{abstract}

\begin{figure*}[t]
    \centering
    \includegraphics[width=1.0\linewidth]{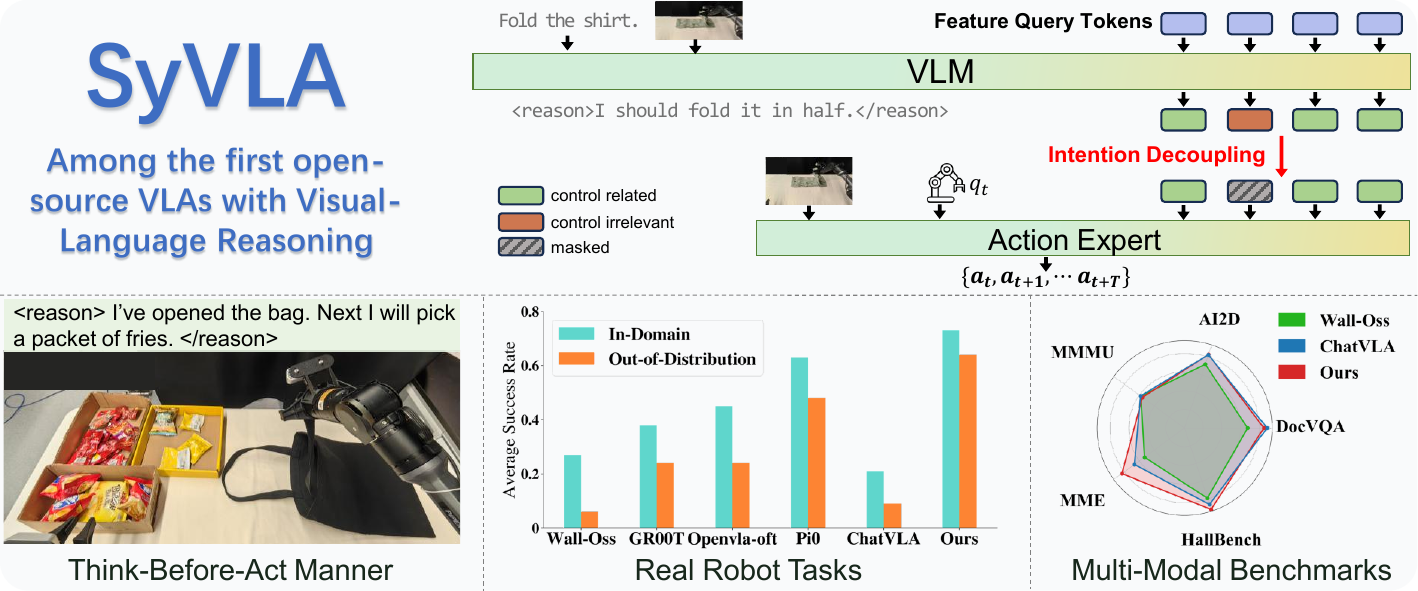}
    \caption{We propose an \textbf{Intention Decoupling} algorithm to disentangle control-irrelevant high-level reasoning information in Feature Query Tokens, and develop an \textbf{RL pipeline} for stable real-world reinforcement learning. With these two methods, our SyVLA model achieves a balance between robotic task competence and visual-language understanding capability preservation.}
    \label{fig:teaser}
\end{figure*}

\section{Introduction}
Driven by large-scale, high-quality robotic datasets collected via expert human teleoperation and the development of Visual-Language Models(VLMs), Vision–Language–Action (VLA) models have advanced rapidly. Several recent representative works ~\cite{black2410pi0, zhai2025igniting, spiritai2026spiritv15, wen2025dvla} have demonstrated that pretraining on over 10k hours of high-quality expert teleoperation data enables highly dexterous manipulation, highlighting the benefits brought by large-scale robotic pretraining.

Nevertheless, two major bottlenecks remain on the path toward human-level general embodied intelligence. First, VLA models often overemphasize action learning, which substantially degrades the inherent visual–language understanding and logical reasoning abilities of the underlying VLM, sometimes even causing catastrophic forgetting. This deviates from human intelligence, where broad task competence emerges from knowledge transfer across domains. Although recent studies ~\cite{zhou2025chatvla, zhouchatvla, zhai2025igniting, intelligence2025pi_} attempt to address this issue via mixed training on multi-modal datasets, we observe that these methods often fail to achieve a satisfactory balance between action competence and vision–language capability. Second, imitation learning(IL) is limited by a mismatch between its training objective and actual task success: during closed-loop control, errors tend to accumulate, gradually driving the agent into out-of-distribution(OoD) observations and ultimately causing failure. Reinforcement learning (RL) is widely considered a promising remedy, as it learns through on-environment rollouts to maintain stable performance. However, the multi-billion-parameter scale of VLA models and the high-dimensional continuous action space make RL training prone to instability, policy drift, and even capability collapse. These two issues constrain VLA models from achieving higher intelligence and more robust performance.

To address these limitations, we propose \textbf{SyVLA}, a new VLA model trained with a three-stage pipeline—pretraining, task fine-tuning, and reinforcement learning. SyVLA adopts a dual-system architecture: a VLM serves as the high-level perception, language-reasoning core and control intention generator; a Flow Matching model serves as the Action Expert, generating actions conditioned on the control intention and multi-modal observations. We connect the two components using a set of \textbf{Feature Query Tokens} that transmit the intention representation. These tokens are appended to the end of the VLM input sequence, and their last hidden states are used as part of the conditioning inputs to the Action Expert. We train on a large-scale robotics dataset, in which only a small fraction (less than 1\%) is annotated with task-oriented chains of thought(CoT), and additionally mix in a certain proportion of multi-modal data. This training strategy enables the VLA model to not only execute robotic tasks in a "Think-Before-Act" manner, but also retain a degree of common-sense knowledge and reasoning ability.

Although SyVLA's Feature Query Tokens are more lightweight than those in other VLA models, we find that the trained SyVLA can easily exhibit imprecise actions or hesitant behavior. We attribute this to the implicit control representation carrying leaked information from higher-level reasoning processes. We therefore propose an annotation-free \textbf{Intention Decoupling} algorithm. During training, we compute the gradient of the action loss with respect to the last hidden state of each Feature Query Token, and mask (by setting to zero) those with the smallest gradient L2 norms, reducing the mutual information between the remaining last hidden states and the reasoning representations, thereby decoupling the implicit control representation. We provide a theoretical insights of the method and experimentally verify that it achieves a balance between improving task performance and preserving vision–language capabilities.

To further unlock SyVLA’s potential, we develop an RL training pipeline based on \textbf{Similar-Sample Guidance}, which stabilizes online RL and mitigates the instability induced by the high-dimensional action space and massive parameter space of VLA models. Specifically, we use similar samples retrieved from the imitation-learning(IL) dataset as update guidance to keep updates close to expert behavior, thereby significantly suppressing policy drift and improving stability and final success rates on long-horizon sparse-reward tasks. By narrowing the gap between the training objective and true task success, the RL stage yields up to a 15\% absolute improvement in success rate over the imitation-initialized policy on long-horizon sparse-reward tasks such as Folding-Shirt.

In summary, our contributions are as follows:

\begin{enumerate}
    \item We propose \textbf{Intention Decoupling} algorithm for VLA training to mitigate the entanglement between control intention and high-level reasoning. Concretely, we mask control-irrelevant last hidden states of Feature Query Tokens to separate control intention representations from reasoning information, leading to more stable real-world task performance. We further provide theoretical insights and validate the effectiveness on real-robot tasks.

    \item We introduce a \textbf{Similar-Sample Guided} RL pipeline. By constructing guidance signals from semantically similar samples in the IL dataset, our method stabilizes RL updates and reduces policy drift.
    
    \item We will open source the pretraining dataset and the full codebase for the entire pipeline. To our knowledge, this is among the first complete open-source VLA releases with visual-language reasoning capability.
\end{enumerate}

\paragraph{Conflict of Interest Disclosure} The authors declare no financial conflicts of interest.
\section{Related Works}
\subsection{VLA Model}
Early VLA models ~\cite{kim2024openvla, zitkovich2023rt, pertsch2025fast} generate discrete action tokens in an auto-regressive manner and decode them into continuous control actions, but this formulation often suffers from limited action precision. In contrast, diffusion policy and flow matching based approaches ~\cite{wen2025diffusionvla, wen2025dexvla, bjorck2025gr00t} synthesize actions through multi-step denoising, enabling finer-grained control and dexterous manipulation; however, they commonly use pretrained VLMs only as initialization, leading to catastrophic forgetting of general vision–language capabilities after training.

Recent works ~\cite{zhouchatvla, zhou2025chatvla, zhai2025igniting, intelligence2025pi_} mix multi-modal data with robotic data to better preserve VLM's inherent ability, enabling them to operate in a “Think-Before-Act” manner. Yet explicit reasoning or subtask decomposition can become entangled with control representations, degrading dexterous performance or causing behavioral stalling. 

Some researchers attempt to alleviate this issue through MoE architectures ~\cite{zhou2025chatvla, zhouchatvla} or intermediate Fast Token ~\cite{pertsch2025fast} representations ~\cite{intelligence2025pi_, zhai2025igniting}, but each comes with its own limitations.

Our approach leverages Feature Query Token and the Intention Decoupling algorithm to effectively preserve general vision–language capabilities, while mitigating the potential coupling between reasoning processes and control representations.

\subsection{Real-Robot RL}
The gap between the training objective of VLA models and the goal of long-horizon task execution often causes failures in closed-loop control due to cumulative error. Consequently, reinforcement learning is essential for improving the task-level performance of VLA models ~\cite{intelligence2025pi}. Yet reinforcement learning on real robotic platforms is unstable and risky.

To mitigate the risks of real-robot exploration, recent work ~\cite{zhu2025wmpo, zhang2025reinforcing, hung2025nora, xiao2025world} has explored using world models as environment simulators to replace unsafe on-robot rollouts and enable safer reinforcement learning. However, these approaches are difficult to extend to deformable manipulation (e.g., clothes folding), where generating high-fidelity observations of deformable objects and fluids remains challenging. $\pi^*_{0.6}$ ~\cite{intelligence2025pi} adopts RECAP algorithm ~\cite{frans2025diffusion}, but it suffers from slow policy optimization and depends on exceptionally strong base models. RL-100 ~\cite{lei2025rl} employs a two-stage pipeline for stable RL training, but it has only been evaluated on small diffusion models, and its effectiveness for VLA-scale models remains to be further validated. Some other researchers also attempt to use methods like residual policies ~\cite{xiao2025self}, human-in-the-loop correction ~\cite{chen2025conrft, luo2024precise}, morphological symmetry augmentation ~\cite{li2025gr}, carefully designed GRPO variants ~\cite{liu2025flow}, etc. However, these approaches heavily rely on human-designed solutions for a certain task, which limits their transferability. Overall, achieving efficient and stable RL optimization for VLA models remains an urgent open problem.

To this end, our Similar-Sample Guided RL method substantially improves the stability of real-world RL for VLA models by retrieving similar samples from IL dataset, thereby enabling a practical RL pipeline for VLAs on real robotic arms.
\section{Methodology}

% In this section, we describe our model architecture, Intention Decoupling algorithm and \textbf{Similar Sample Contrastive RL pipeline} in detail. In Section~\ref{sec:3.1}, we first introduce our \textbf{SyVLA} model, which is a dual-system-structured VLA model, and effectively preserves the inherent language and reasoning capabilities of the VLM model. Section~\ref{sec:3.2} details our newly proposed \textbf{Intention Decoupling} algorithm, which can substantially alleviate the entanglement between the SyVLA model’s high-level reasoning process and implicit control intentions, thereby improving SyVLA's performance. We also theoretically prove the effectiveness of the algorithm. In Section~\ref{sec:3.3}, we describe in detail our real-world RL pipeline based on Similar-Sample Contrast, which can significantly enhance the stability of RL training and yield considerable performance improvements.

In this section, we describe our model architecture, \textbf{Intention Decoupling algorithm} and \textbf{Similar Sample Guided RL pipeline} in detail. We further offer theoretical insights of the effectiveness of our algorithm at the end of Section~\ref{sec:3.2}

\subsection{SyVLA}\label{sec:3.1}

\begin{figure*}[t]
    \centering
     \includegraphics[width=1.0\linewidth]{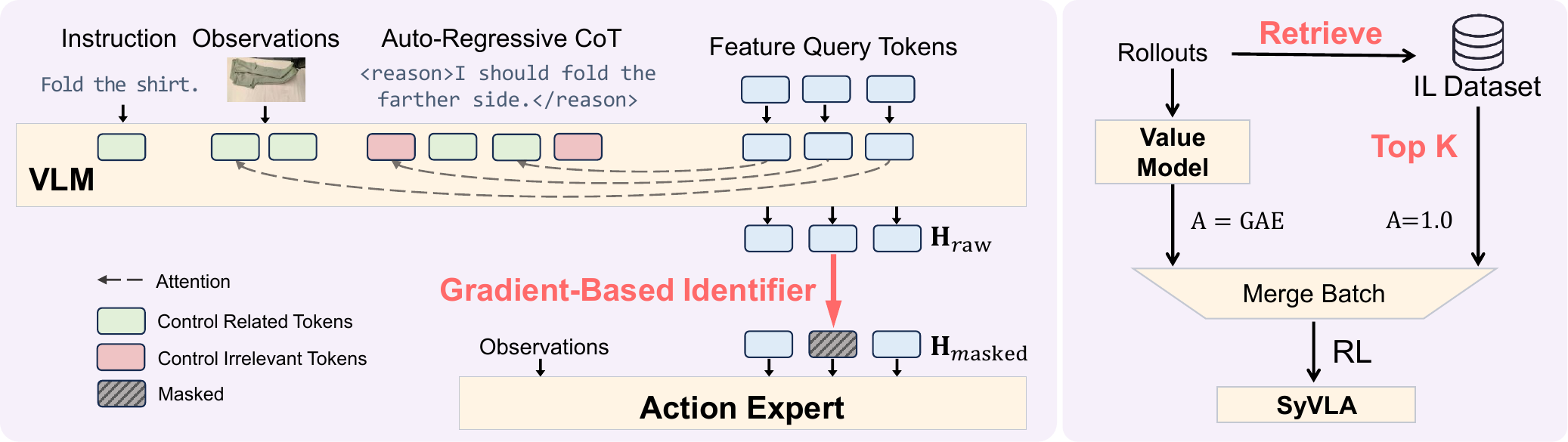}
     \caption{\textbf{Overview of our Intention Decoupling algorithm and our Similar Sample Guided RL pipeline.} \textbf{Left}: To avoid degradation in the SyVLA’s action capability, we use a gradient-based identifier to find the tokens weakly associated with the control intention representation, and then mask these tokens' last hidden state before feeding them into the Action Expert. \textbf{Right}: To stabilize our RL training, we select samples semantically similar to the current rollout samples from IL dataset and merge them into a batch for training. We set the advantage of samples from IL dataset to 1.0 for stable optimization.}
    \label{fig:syvla}
\end{figure*}

As shown in Figure~\ref{fig:syvla}, our model is divided into two components: a VLM and an Action Expert. We use Qwen2.5VL-3B ~\cite{bai2025qwen2} as our VLM, and equip it with a transformer-based Flow Matching model ~\cite{lipman2022flow} as our Action Expert. Unlike Pi0’s KV Cache ~\cite{black2410pi0}, we choose to use a fixed set of learnable Feature Query Tokens to connect the two components of the SyVLA model. We append the Feature Query Tokens after the VLM’s auto-regressive chain-of-thought(CoT) output and use their last hidden states(hereafter, \textbf{Feature Query States}) as part of the input to the Action Expert. In the Action Expert, the Feature Query States are first passed through an MLP adapter to produce the control intention representation condition for flow matching, which guides the Action Expert to generate actions consistent with the VLM’s plan. Compared with the KV cache approach adopted in Pi0, our Feature Query Token approach is more lightweight, leading to lower inference latency. Moreover, it can be readily extended to support asynchronous inference between the VLM and the Action Expert, maximizing the model’s inference efficiency.

 % We emphasize that this is consistent with the structure of the human nervous system: the VLM viewed as the human brain is responsible for high-level cognitive activities and for generating action intentions, while the Action Expert viewed as the spinal cord is responsible for coordinating the motion of joints and muscles. In this setting, the Feature Query Tokens play the role of neural signal transmission in the brain stem. Through a limited information-transmission bottleneck, the VLM is forced to generate more essential control-intention signals, rather than the high-level signals related to cognition, thinking, and decision-making, which the Action Expert does not care about. Some may notice that we still use the Siglip model ~\cite{zhai2023sigmoid} to encode visual signals in the Action Expert. It stems from a key insight into human behavioral mechanisms: in some cases, humans can rapidly produce reflexive responses without deliberate thinking. This justifies bypassing part of the visual signals around the brain’s decision-making and directly transmitting them to the Action Expert. Meanwhile, we emphasize that, compared with the KV cache approach adopted in Pi0, our Feature Query Token approach is more lightweight, leading to lower inference latency. Moreover, it can be readily extended to support asynchronous inference between the VLM and the Action Expert, thereby further maximizing the model’s inference efficiency.

Many Previous works ~\cite{black2410pi0, wen2025diffusionvla, kim2025fine, bjorck2025gr00t} primarily leverage VLM’s powerful visual understanding capability while neglecting its more important language capability, often leading to catastrophic forgetting after training. We argue this is misguided since humans’ broad generalization ability largely attributes to vision–language reasoning capability. Accordingly, SyVLA is trained not only on action datasets but also jointly on several high-quality multi-modal datasets, preserving its inherent capabilities while improving spatial cognition. Benefiting from this training paradigm, our model preserves visual-language understanding capability to a large extent. Moreover, SyVLA is further trained to leverage an explicit reasoning process to decompose subtasks, which enables it to solve more tasks requiring high-level thinking and were out of reach for prior VLA models.

During inference, we first allow the VLM to perform limited reasoning to handle potentially abstract user instructions or complex situations, and then append the Feature Query Tokens after the generated reasoning tokens to get Feature Query States, which are used to guide the Action Expert to generate corresponding actions according to VLM's plan.

More implementation details about SyVLA and training pipeline can be found in the Appendix~\ref{app:implement}.

\subsection{Intention Decoupling Algorithm}\label{sec:3.2}
Although the training pipeline and the “Think-Before-Act” manner introduced above greatly expand SyVLA’s task capability boundary and generalization ability, we find that our SyVLA model can be prone to poor action precision and indecisiveness in some cases. We attribute this to the leakage of the higher-level reasoning process, which causes the Action Expert confused by the high-level thoughts and wobbled between different decisions.

% This can be attributed to the leakage of the higher-level reasoning process, which causes the Action Expert confused by the high-level thoughts and wobbled between different decisions.

To address this issue, we conducted a theoretical analysis and, guided by the resulting theoretical insights, developed a gradient-based \textbf{Intention Decoupling} algorithm, which suppresses redundant information in the last hidden states of Feature Query Tokens (\textbf{Feature Query States}), thereby enhancing the informational effectiveness of the intention condition used to guide the Action Expert.

We adopt a two-step forward procedure to mask invalid information. In the first step, we feed the Feature Query States
\[
\mathbf{H}_{\mathrm{raw}} = \{ \mathbf{h}_{\mathrm{raw}}^{0}, \mathbf{h}_{\mathrm{raw}}^{1}, \ldots, \mathbf{h}_{\mathrm{raw}}^{n-1} \}
\]
produced by the VLM, together with other required inputs, into the Action Expert to predict and compute the action loss \(L_{\mathrm{action}}\) once. Then, we obtain its gradients with respect to the hidden states $\mathbf{H}_{\mathrm{raw}}$,
\[
\mathbf{G} = \{ \mathbf{g}^{0}, \mathbf{g}^{1}, \ldots, \mathbf{g}^{n-1} \},
\qquad
\mathbf{g}^{i} = \frac{\partial L_{\mathrm{action}}}{\partial \mathbf{h}_{\mathrm{raw}}^{i}}
\]

Our theoretical analysis demonstrates that the hidden states with small corresponding gradient norms contribute little to control intention representation and mainly contain redundant information. Accordingly, we compute the \(\ell_{2}\)-norm of each gradient vector \(\lVert \mathbf{g}^{i} \rVert_{2}\) and mask the Feature Query States whose \(\ell_{2}\)-norm value fall below a threshold. This threshold is empirically chosen as the 5th percentile of the gradient-norm distribution, which can deliver strong performance across multiple tasks. The $i$-th Feature Query State after masking is then given by
\[
\mathbf{h}_{\mathrm{masked}}^{i} = \mathbf{h}_{\mathrm{raw}}^{i} \cdot \mathbb{I}\left( \lVert \mathbf{g}^{i} \rVert_{2} \ge \tau \right), i = 0, 1, \cdots n-1
\]
where \(\tau\) denotes the selected \(q\)-quantile threshold and \(\mathbb{I}(\cdot)\) is the indicator function. We then perform a second forward pass using masked Feature Query States \(\mathbf{H}_{\mathrm{masked}}\), recompute \(L_{\mathrm{action}}\), and proceed with subsequent computations and model updates.

Some may argue that our method introduces an additional forward-backward pass and thus incurs a high computational burden. We clarify that the additional computation of our method only involves the Action Expert component, which typically accounts for less than 20\% of the total parameters of the VLA model, and our method does not involve additional model updates. Therefore, it does not lead to a significant degradation in training speed. In practice, our training time increases by only about 10\%, while yielding substantial performance improvements.

We further show the effectiveness of our method through theoretical analysis; the key results are summarized as follows.

We simplify our Action Expert model as a single layer transformer. Under this circumstance, we have the following theorem.
\begin{theorem}
The Action Expert module can be denoted as
$$
    \hat{a} = \psi(z) = \psi \left( \frac{\sum_{i=1}^n \kappa(q, W^Kh_i) W^Vh_i}{\sum_{i=1}^n \kappa(q, W^Kh_i)} \right)
$$
, where $W^k, W^v$ are the K, V mapping matrices of the self-attention layer, $h_i$ is the $i$-th Feature Query State, $q$ is a learnable query vector, $z$ is the output of self-attention layer, $\hat{a}$ is the predicted action, $\psi$ is a non-linear function.
\end{theorem}

\paragraph{Remark} The above theorem denotes a single layer Action Expert as a function mapping input Feature Query States $\mathbf{H}$ to the predicted action $\hat{a}$. 

\begin{theorem}
    The gradient of loss function $\ell$ regarding to the $i$-th Feature Query State is
    $$
        \frac{\partial \ell}{\partial h_i} = \frac{\partial \ell}{\partial z} \cdot \frac{\kappa(q, W^K h_i)}{\sum_{j=1}^n \kappa(q, W^K h_j)} \cdot \left[ \left( v_i - z \right) \cdot q^T W^k + W^v \right]
    $$
    , where $v_i = W^V h_i$ is the $i$-th value vector, and $\frac{\kappa(q, W^K h_i)}{\sum_{j=1}^n \kappa(q, W^K h_j)}$ is the $i$-th normalized attention score.
\end{theorem}

\paragraph{Remark} Since $W^k, W_v, q$ are the same for all input Feature Query States $h_i$, the above theorem demonstrates that the value of $\frac{\partial \ell}{\partial h_i}$ is determined by two factors: the magnitude of the attention score, and the distance between \(v_i\) and the output of attention layer $z$.

For the first case, a small attention score indicates that the model considers the \(i\)-th input to be irrelevant to the decision.

For the second case, when \(v_i\) is close to the attention layer’s output \(z\), it indicates that the information in \(v_i\) is largely already covered by other Feature Query Tokens, and thus can be regarded as redundant. However, keeping \(v_i\) for decision-making may cause the model to learn a shortcut decision path and ignore the true causal relevance. This could be acceptable under in-domain cases since it doesn't harm the output, while in OoD cases this could lead to non-causal decision process and degraded performance.

A more detailed proof of the theorem and more analysis can be found in Appendix~\ref{app:theorem}.

\subsection{Similar-Sample Guided RL Pipeline}\label{sec:3.3}
In addition to the advantages introduced above, we observe that our method also provides a better starting point for reinforcement learning of VLA models.

VLA models trained solely with imitation learning often lack robustness. This is because the loss function during imitation learning encourages the model to exactly fit expert actions at each time step, whereas real-task success is jointly determined by the closed-loop execution of over hundreds of actions. This creates a significant gap between the training objective and the real-world task success rate. During real-world execution, small action errors can gradually accumulate and induce distribution shift, driving the robot’s observations out of the training distribution (OoD) and ultimately causing task failure. Therefore, we emphasize the necessity of performing reinforcement learning for VLA in the real world.%: through closed-loop interaction and reward optimization, the model can learn to self-correct based on feedback, improving VLA's robustness.

However, due to their high-dimensional, open-ended perceptual space and the high cost of real-robot exploration, VLA models are highly prone to performance collapse or make risky rollouts, especially for tasks requiring fine manipulation. We observe our Intention Decoupling algorithm improves the model’s generalization ability, substantially enhancing its capability to handle OoD situations. % In practice, using only all-success data, it can exhibit partial error-recovery ability, which provides reinforcement learning with an excellent initial policy and greatly improves rollout efficiency.

To further stabilize the reinforcement learning process, we also develop an RL pipeline that leverages similar samples from the expert dataset collected for imitation learning. We encode the observation at each timestep into a feature vector. At each reinforcement learning update, in addition to sampling data from the rollout dataset, we retrieve similar data from the expert dataset and merge all samples into a batch for policy update. This training scheme allows the policy to correct stepwise action deviations during updates while preserving overall behavioral consistency. Specifically, we use the following formula to compute the similarity between sample x and sample y:

\[
\mathrm{sim}(x,y)\;=\;\sum_{v \in \mathcal V} w_v \cdot \mathrm{sim_{cos}}\!\left(E\!\left(O^{(v)}_x\right),\,E\!\left(O^{(v)}_y\right)\right)
\]
,where $O_x^{(v)}$ and $O_y^{(v)}$ denote the images of samples \(x\) and \(y\) observed from view \(v\in\mathcal V\), \(E(\cdot)\) is an image encoder, \(\mathrm{sim_{cos}}(\cdot,\cdot)\) is the cosine similarity, and \(w_v\ge 0\) is the weight for view \(v\).

Nevertheless, we find that applying standard policy-gradient updates to samples from the expert dataset leads to rapid divergence—even when we completely remove rollout data: both the loss and gradient norm explode within 1k updates. We hypothesize that this is caused by a severe mismatch between the value model estimation and the expert behavior distribution, resulting in high-variance and misleading advantage estimates. Therefore, we set the advantage of samples from the expert dataset uniformly to 1, making the training objective equivalent to maximizing expert action likelihood. In practice, we find that this stabilizes VLA policy updates to the greatest extent, keeping the updated actions reasonable and largely avoiding the policy capability collapse commonly observed in reinforcement learning. With this strategy, we obtain a stable real-world reinforcement learning pipeline, further improving the robustness and performance of our VLA model.

\begin{figure*}[t]
    \centering
    \includegraphics[width=1.0\linewidth]{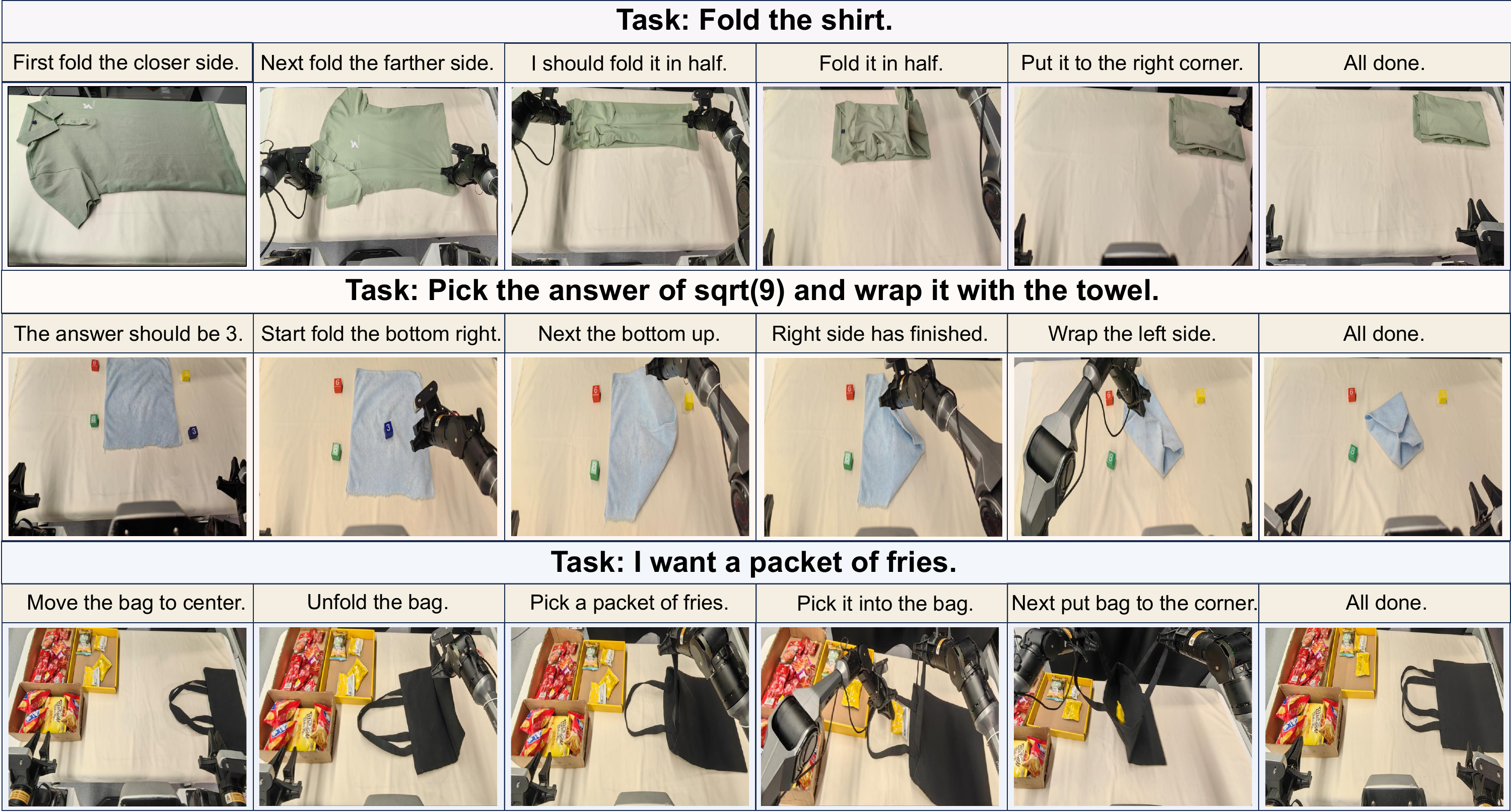}
    \caption{\textbf{Visualization of three tasks.} The figure presents the scenes of three real-world robotic tasks and illustrates the process by which our SyVLA completes these tasks in a “Think-Before-Act” manner.}
    \label{fig:task}
\end{figure*}

More implementation details and design motivations can be found in Appendix~\ref{app:implement}.
\section{Experiments}

% In this section, we systematically evaluate the effectiveness and generalization capability of the proposed model and method across multiple representative tasks. We first introduce the model configuration, training procedure, and evaluation settings used in our experiments. We then conduct comparative experiments against recent strong SOTA methods on challenging real-world tasks as well as commonly used VQA benchmarks. Finally, we analyze the impact of key design choices on performance through ablation studies. The results show that our method achieves consistent improvements across multiple settings and demonstrates stronger generalization capability.

\subsection{Experiment Settings}
\subsubsection{Evaluation Tasks}\label{sec:4.1.1}
We conduct experiments on our model and compare with several strong baselines in two domains: real-world tasks and commonly used multi-modal benchmarks. 

For real-world tasks, we conduct our experiments on the Cobot Magic robotic platform, reporting task success rate as the metric. To better validate the generalization ability of VLA models, we design a task-specific out-of-distribution(OoD) setting for each task. Our tasks are shown in Figure~\ref{fig:task}, and are introduced as follows.

% \begin{figure*}[t]
%     \centering
%     \includegraphics[width=1.0\linewidth]{figs/tasks.pdf}
%     \caption{\textbf{Visualization of three tasks.} The figure presents the scenes of three real-world robotic tasks and illustrates the process by which our SyVLA completes these tasks in a “Think-Before-Act” manner.}
%     \label{fig:task}
% \end{figure*}

\textbf{Task 1 Folding Shirts}. The VLA is asked to fold a Polo shirt. This task requires the VLA model to perform multi-step, fine-grained manipulations. In the OoD setting, we place the shirt at an off-center, previously unseen location to challenge the model’s positional generalization. For this task, each model conducts 14 trials.

\textbf{Task 2 Calculating and Wrapping}. Given a random arithmetic instruction asked by the user, select the cube with the correct answer and wrap it with a towel. This task not only requires the model to possess basic arithmetic knowledge, but also tests its fine manipulation ability. In the OoD setting, we provide instructions never seen during training to verify that success comes from intrinsic reasoning ability rather than simple memorization of instruction-to-action mappings. For this task, each model conducts 28 trials.

\textbf{Task 3 Bagging Snacks}. Given a vague instruction from the user, select the correct snack and place it into an initially folded cloth bag. This task is highly practical and indicates the potential for deploying VLA models in future unmanned retail. In the OoD setting, we likewise provide vague instructions never seen during training. For this task, each model conducts 14 trials.

In addition to these real-world tasks, we also select multiple representative multi-modal benchmarks, including DocVQA ~\cite{mathew2021docvqa}, AI2D ~\cite{kembhavi2016diagram}, MMMU ~\cite{yue2024mmmu}, MME ~\cite{fu2025mme}, and HallBench ~\cite{guan2024hallusionbench}, to verify whether the VLM component still retains its original knowledge and reasoning capabilities.

\subsubsection{Implementation Details of SyVLA}
We choose Qwen2.5VL-3B ~\cite{bai2025qwen2} as the VLM backbone of our SyVLA model and build an Action Expert based on a Transformer Flow Matching model. For the number of Feature Query Tokens, we set n=20, which performs sufficiently well across all our tasks.

Our training procedure follows a three-stage pipeline. First, we pretrain our VLA model on a large-scale robotic dataset, mixed with a multi-modal dataset at an approximate ratio of 30\%. Then, in the second stage, we collect several hundred trajectories for each target task and train the pretrained SyVLA to acquire task-specific solutions. After the second stage, the model is already able to complete the tasks with a relatively good success rate. Finally, to maximize the performance of the VLA model, we adopt the reinforcement learning pipeline described in Section~\ref{sec:3.3} and adapt PPO~\cite{schulman2017proximal} to fit our RL algorithm to optimize our VLA model in the real world. We emphasize that our Intention Decoupling algorithm described in Section~\ref{sec:3.2} is applied throughout the entire model training process.

% First, we pretrain our VLA model on a large-scale action dataset, which mainly includes a subset of the open-source dataset Open X Embodiment ~\cite{open_x_embodiment_rt_x_2023}, as well as an approximately 10h human teleoperation dataset collected by us. This stage is primarily used to train the randomly initialized Action Expert model and the Feature Query Tokens. Throughout pretraining, we consistently mix multi-modal datasets at an approximately 30\% ratio to prevent catastrophic forgetting in the VLM model. Then, in the second stage, we collect several hundred task-specific trajectories and train the pretrained SyVLA model to acquire task-specific solutions. After this step, the VLA model can already complete the entire task with a relatively good success rate. Finally, to maximize the performance of our SyVLA model, we adopt the online reinforcement learning pipeline described in Section ~\ref{sec:3.3} and choose PPO as our base reinforcement learning algorithm to optimize our model in the real world. In this stage, we allow up to 10 rollout-update iterations, where each rollout data contains 7 episodes. After three-stage training, our model can perform real-world tasks well. We emphasize that our Intention Decoupling algorithm described in Section ~\ref{sec:3.2} is applied throughout the entire model training process.

More implementation details can be found in Appendix~\ref{app:implement}.

% \begin{table*}[t]
%     \centering
%     \begin{tabular}{ccccccccc}
%     \toprule
%     Method
%     & \multicolumn{4}{c}{In Domain}
%     & \multicolumn{4}{c}{Our of Distribution} \\
%     \cmidrule(lr){2-5}\cmidrule(lr){6-9}
%     & Task 1 & Task 2 & Task 3 & Avg. Success Rate
%     & Task 1 & Task 2 & Task 3 & Avg. Success Rate \\
%     \midrule
%     OpenVLA-oft        & 0.71 & 0.29 & 0.36 & 0.45 & 0.55 & 0.11 & 0.07 & 0.24 \\
%     GR00T              & 0.71 & 0.21 & 0.21 & 0.38 & 0.64 & 0.07 & 0.00 & 0.24 \\
%     Wall-Oss           & 0.50 & 0.18 & 0.14 & 0.27 & 0.14 & 0.04 & 0.00 & 0.06 \\
%     Pi0 (pretrained)   & \textcolor{red}{0.93} & 0.39 & 0.57 & 0.63 & \textcolor{red}{0.78} & 0.29 & 0.36 & 0.48 \\
%     Pi0 (from scratch) & 0.64 & 0.21 & 0.29 & 0.38 & 0.50 & 0.14 & 0.14 & 0.26 \\
%     ChatVLA            & 0.21 & 0.29 & 0.14 & 0.21 & 0.00 & 0.21 & 0.07 & 0.09 \\
%     \midrule
%     SyVLA (ours)       & 0.86 & \textcolor{red}{0.68} & \textcolor{red}{0.64} & \textcolor{red}{0.73} & \textcolor{red}{0.78} & \textcolor{red}{0.57} & \textcolor{red}{0.57} & \textcolor{red}{0.64} \\
%     \bottomrule
%     \end{tabular}
%     \caption{\textbf{Results on real world tasks.} We evaluate multiple models under both In Domain and OoD settings. The experimental results show that our SyVLA significantly outperforms other strong baselines and exhibits better generalization ability. This demonstrates the effectiveness of our method.}
%     \label{tab:main_exp}
% \end{table*}

\begin{table*}[t]
    \centering
    \begin{tabular}{ccccccccc}
    \toprule
    Method
    & \multicolumn{4}{c}{In Domain}
    & \multicolumn{4}{c}{Out of Distribution} \\
    \cmidrule(lr){2-5}\cmidrule(lr){6-9}
    & Task 1 & Task 2 & Task 3 & Avg. Success Rate
    & Task 1 & Task 2 & Task 3 & Avg. Success Rate \\
    \midrule
    OpenVLA-oft        & 0.71 & 0.29 & 0.36 & 0.45 & 0.55 & 0.11 & 0.07 & 0.24 \\
    GR00T              & 0.71 & 0.21 & 0.21 & 0.38 & 0.64 & 0.07 & 0.00 & 0.24 \\
    Wall-Oss           & 0.50 & 0.18 & 0.14 & 0.27 & 0.14 & 0.04 & 0.00 & 0.06 \\
    Pi0 (pretrained)   & \textbf{0.93} & 0.39 & 0.57 & 0.63 & \textbf{0.78} & 0.29 & 0.36 & 0.48 \\
    Pi0 (from scratch) & 0.64 & 0.21 & 0.29 & 0.38 & 0.50 & 0.14 & 0.14 & 0.26 \\
    ChatVLA            & 0.21 & 0.29 & 0.14 & 0.21 & 0.00 & 0.21 & 0.07 & 0.09 \\
    \midrule
    SyVLA (ours)       & 0.86 & \textbf{0.68} & \textbf{0.64} & \textbf{0.73} & \textbf{0.78} & \textbf{0.57} & \textbf{0.57} & \textbf{0.64} \\
    \bottomrule
    \end{tabular}
    \caption{\textbf{Results on real world tasks.} We evaluate multiple models under both In Domain and OoD settings. The experimental results show that our SyVLA significantly outperforms other strong baselines and exhibits better generalization ability. This demonstrates the effectiveness of our method.}
    \label{tab:main_exp}
\end{table*}

% \begin{table*}[h]
%     \centering
%     \begin{tabular}{c|ccccc}
%     \toprule
%     & DocVQA & AI2D & MMMU & MME & HallBench \\
%     \midrule
%     OpenVLA-oft & - & - & - & - & - \\
%     Wall-Oss & TBD & TBD & TBD & TBD & TBD \\
%     GR00T & - & - & - & - & - \\
%     Pi0 & - & - & - & - & - \\
%     ChatVLA & \textcolor{red}{83.3} & 67.36 & \textcolor{red}{37.4} & 1435 & 39.90 \\
%     \midrule
%     Ours & 80.01 & \textcolor{red}{67.6} & 35.78 & \textcolor{red}{1795} & \textcolor{red}{42.53} \\
%     \bottomrule
%     \end{tabular}
%     \caption{\textbf{Results on VQA Benchmarks.} We evaluate our SyVLA and other baselines on multiple representative VQA benchmark tasks. Our model outperforms other baselines in commonsense knowledge and visual detail understanding, while retaining comparable performance on capabilities such as document understanding and multidisciplinary knowledge.}
%     \label{tab:vqa_exp}
% \end{table*}

\begin{table*}[t]
    \centering

    \raisebox{-0.8\height}{
        \begin{minipage}[t]{0.66\textwidth}
            \centering
            \begin{tabular}{c|ccccc}
            \toprule
            & DocVQA & AI2D & MMMU & MME & HallBench \\
            \midrule
            OpenVLA-oft & - & - & - & - & - \\
            Wall-Oss & 63.62 & 58.60 & 37.11 & 1146.56 & 36.57 \\
            GR00T & - & - & - & - & - \\
            Pi0 & - & - & - & - & - \\
            ChatVLA & \textbf{83.30} & \underline{67.36} & \textbf{37.40} & \underline{1435} & \underline{39.90} \\
            \midrule
            Ours & \underline{80.01} & \textbf{67.70} & \underline{35.78} & \textbf{1795} & \textbf{42.53} \\
            \bottomrule
            \end{tabular}
        \end{minipage}
        \hfill
        \begin{minipage}[t]{0.34\textwidth}
        \centering
        \raisebox{-0.45\height}{
            \includegraphics[width=\linewidth]{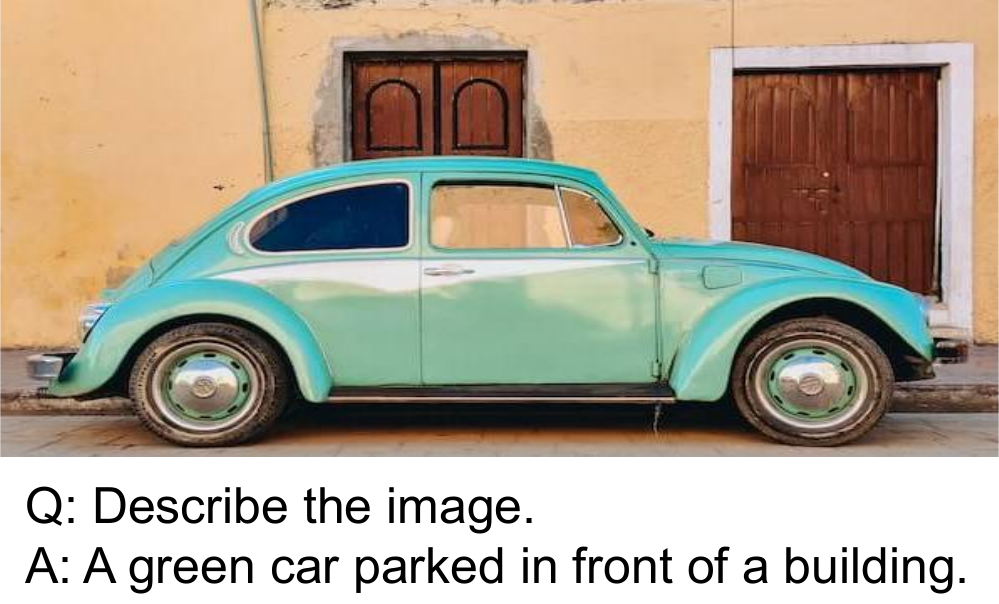}
        }
        \end{minipage}
    }

    \caption{\textbf{Results on multi-modal benchmarks.} We evaluate our SyVLA and other baselines on multiple representative multi-modal benchmarks. Our model outperforms other baselines in commonsense knowledge and visual detail understanding, while retaining comparable performance on capabilities such as document understanding and multidisciplinary knowledge. We use - to indicate the VLA model has no VQA capability. \textbf{Right} is a Visual QA example of our SyVLA.}
    \label{tab:vqa_exp}
\end{table*}

\subsubsection{Baselines}
We compare our method with multiple representative strong baselines, including Pi0 ~\cite{black2410pi0}, ChatVLA ~\cite{zhou2025chatvla}, OpenVLA-oft ~\cite{kim2025fine}, Wall-Oss ~\cite{zhai2025igniting}, and GR00T ~\cite{bjorck2025gr00t}, and finetune them using our teleoperation datasets. However, we note that the previous SOTA model Pi0 is pretrained on a proprietary dataset of over 10k hours, which we have no access to. For a fairer comparison, we train a Pi0 model from scratch based on the pretrained VLM on SyVLA’s training datasets to factor out the performance gains brought by large-scale pretraining and potential data leakage. In the following, we use Pi0(from scratch) to denote the Pi0 model trained by us from scratch, and Pi0(pretrained) to denote the Pi0 model finetuned from the official pretrained checkpoint. Detailed baseline introduction can be found in Appendix~\ref{app:baseline}.

\subsection{Main Results}

% As shown in Table ~\ref{tab:main_exp}, we evaluated our SyVLA against the strong baselines on the real-world tasks. Evidently, SyVLA achieves the best performance on two of the tasks and significantly outperforms the baselines. For Task 1, SyVLA performs slightly worse than the pretrained Pi0 model, yet substantially better than Pi0 trained on the same dataset. This indicates that a large portion of Pi0’s capabilities stem from its massive pretraining data rather than from the model architecture and training algorithm per se. In contrast, our SyVLA requires less than 5\% of the pretraining data used by official Pi0 to attain performance close to that of the large-scale pretrained Pi0 model, thereby demonstrating the effectiveness of our approach.

As shown in Table~\ref{tab:main_exp}, we evaluate our SyVLA against several strong baselines on the real-world tasks.

In the in-domain setting, our model achieves the best performance on two of the three tasks, significantly outperforming the other baselines. For Task 1, SyVLA performs slightly worse than the pretrained Pi0 model, yet substantially better than Pi0 model trained on the same dataset. This indicates that the advantage of Pi0 mostly stems from its massive pretraining data rather than its architecture and training strategy per se. In contrast, our SyVLA uses less than 5\% of pi0’s pretraining data while attaining performance close to that of the large-scale pretrained pi0 model, strongly demonstrating the effectiveness of our approach.  

In the out-of-distribution setting, our method significantly outperforms all baselines and suffers from the smallest performance drop. This suggests that our method has superior generalization ability and shows promise in understanding complex user instructions.

Notably, ChatVLA adopts an MoE architecture to mitigate gradient conflicts arising from joint training on action data and multi-modal data, thereby avoiding catastrophic forgetting. However, we observe that it suffers from a severe degradation in fine-grained manipulation ability: while it can follow user instructions to select the correct target object, it fails to execute subsequent precise manipulations.

% \begin{table*}[h]
%     \centering
%     \begin{tabular}{c|ccccc}
%     \toprule
%     & DocVQA & AI2D & MMMU & MME & HallBench \\
%     \midrule
%     OpenVLA-oft & - & - & - & - & - \\
%     Wall-Oss & TBD & TBD & TBD & TBD & TBD \\
%     GR00T & - & - & - & - & - \\
%     Pi0 & - & - & - & - & - \\
%     ChatVLA & \textcolor{red}{83.3} & 67.36 & \textcolor{red}{37.4} & 1435 & 39.90 \\
%     \midrule
%     Ours & 80.01 & \textcolor{red}{67.6} & 35.78 & \textcolor{red}{1795} & \textcolor{red}{42.53} \\
%     \bottomrule
%     \end{tabular}
%     \caption{\textbf{Results on VQA Benchmarks.} We evaluate our SyVLA and other baselines on multiple representative VQA benchmark tasks. Our model outperforms other baselines in commonsense knowledge and visual detail understanding, while retaining comparable performance on capabilities such as document understanding and multidisciplinary knowledge.}
%     \label{tab:vqa_exp}
% \end{table*}

Beyond real-world tasks, we also evaluated our model’s multi-modal capabilities on several commonly used benchmarks. 
% We report results using the Stage-2–trained SyVLA model, because the goal of Stage-3 is to specialize the model for a given task rather than to maintain a general-purpose capability; therefore, some loss in multi-modal understanding is acceptable. 
The results are shown in Table~\ref{tab:vqa_exp}. 

SyVLA achieves the best performance on AI2D, MME, and HallBench, while remaining only slightly behind ChatVLA on DocVQA and MMMU. We argue that this outcome is intuitive: the primary objective of VLA models is to execute user-issued tasks, which requires strong capabilities in fine-grained visual understanding and broad common knowledge. To endow the VLM with task decomposition and action planning abilities, our SyVLA exhibits a larger decline in document understanding and multidisciplinary expert reasoning capabilities, which are less directly related to VLA tasks. This trade-off is an unavoidable cost given limited resources of data. In comparison, ChatVLA delivers strong results across all benchmarks but is nearly unable to accomplish highly dexterous real-world tasks, which we attribute to its design choice to prioritize retaining vision–language capabilities. Pi0 represents the opposite extreme: although it acquires broad task-execution ability via large-scale pretraining, it completely loses the inherent vision–language understanding ability of its VLM. Overall, SyVLA achieves the best balance between robotic task competence and vision–language understanding capability preservation.

% Compared with ChatVLA, our method demonstrates a clear advantage, outperforming ChatVLA2 on all benchmarks. Relative to the original Qwen2.5VL-3B model, our method incurs less than 10\% performance drop, indicating the superiority of the SyVLA architecture and the effectiveness of our training strategy.

Taken together, these two aspects of experiments show that, compared with other recent competitive baseline models, our SyVLA achieves strong performance on both real-world tasks and multi-modal understanding benchmarks. We further emphasize that SyVLA is pretrained with an extremely limited dataset, which amounts to less than 5\% of the pretraining data used by Pi0 and is also substantially smaller than that used by Wall-Oss and GR00T. This provides additional evidence for the effectiveness and practical value of our architectural choices and training-algorithm design.

\subsection{Ablation Studies}
To further validate the effectiveness of our Intention Decoupling algorithm and the Similar-Sample Guided RL pipeline, we conduct a detailed ablation study. Experiments are performed on \textbf{Task 1 Folding Shirt} since this task challenges the model’s long-horizon, fine-grained execution ability.

We evaluate multiple settings and compare them with the full model, \textbf{SyVLA (all)}, as follows:

\textbf{w/o CoT} During 3-stage training, We disable the "Think-Before-Act" manner of our model and also do not use our Intention Decoupling algorithm. This is intended to verify our earlier claim that the degradation in SyVLA’s performance should be attributed to the leakage of its high-level reasoning process.

\textbf{w/o Intention Decoupling}: We remove Intention Decoupling algorithm while keeping the three-stage training pipeline unchanged.

\textbf{w/o RL}: We disable the RL Stage. For a fair comparison, we collect (via teleoperation) a dataset with the same number of episodes as in RL and continue training using imitation learning.

\textbf{w/o Expert Dataset} During the RL stage, we do not use the expert dataset and train solely on the rollout dataset.

\textbf{w/o Similar Sample} We do not perform similar sample retrieval during RL. Instead, we randomly sample data from the expert dataset.

\textbf{w/ standard Advantage} For samples from the expert dataset, we follow PPO and compute standard GAE for policy-gradient updates.

\begin{table}[h]
    \centering
    \begin{tabular}{lc}
       \toprule
       Setting & Average Success Rate \\
       \midrule
       \textbf{w/o} CoT                  & 0.79 \\
       \textbf{w/o} Intention Decoupling & 0.43 \\
       \textbf{w/o} RL                   & 0.71 \\
       \textbf{w/o} Expert Dataset       & 0.21 \\
       \textbf{w/o} Similar Sample       & 0.79 \\
       \textbf{w/} Standard Advantage    & 0.00 \\
       \midrule
       SyVLA(all)                        & 0.86 \\
       \bottomrule
    \end{tabular}
    \caption{\textbf{Ablation Study}. We conduct ablation study under several settings to verify the effectiveness of our key designs. The results demonstrate that all our designs are essential for the performance.}
    \label{tab:ablation}
\end{table}

As shown in Table~\ref{tab:ablation}, the result of \textbf{w/o CoT} validates our earlier hypothesis that the performance degradation stems from the coupling between high-level reasoning information and control intentions.
With Think-Before-Act enabled, the \textbf{Intention Decoupling} algorithm effectively addresses this issue and further delivers superior performance.

We also verify the necessity of reinforcement learning. It bridges the gap between the training objective and task success, preventing observation OoD drift caused by the accumulation of errors. Under \textbf{w/o Expert Dataset}, we find the RL updates to be highly unstable: after a single update, the model can easily acquire different undesirable behaviors, or even suffer a substantial performance drop. While simply introducing Expert Dataset already brings a marked stabilizing effect, we find that \textbf{Similar Sample} tends to offer a higher performance upper bound.

Notably, using standard Generalized Advantage Estimation(GAE) on the expert dataset to guide policy-gradient updates consistently leads to rapid gradient explosion. We attribute this issue to incorrect advantage assignment, which instead confuses the update direction of the VLA model and results in capability collapse.

Overall, the ablation study provides strong evidence for the effectiveness of our designs. 

\subsection{Hyperparameter Sensitivity}
To examine the sensitivity of our method to hyper-parameters, we conduct ablation studies on the mask threshold $\tau$ and the number of Feature Query Tokens $N$. In addition, we conduct with 5 different random seeds on the 3 tasks in Table~\ref{tab:main_exp} and report the standard deviation of the success rate. The results are as follows.

\begin{table}[h]
    \centering
    \begin{tabular}{cccccccc}
       \toprule
       $\tau$ & 0\% & 3\% & 5\% & 8\% & 10\% & 15\% & 20\% \\
       \midrule
              & 0.43 & 0.73 & 0.86 & 0.77 & 0.86 & 0.82 & 0.64 \\
       \bottomrule
    \end{tabular}
    \caption{\textbf{Sensitivity of $\tau$}. We conduct ablation study with different masking threshold percentage of Intention Decoupling algorithm. The performance remains relatively stable when $\tau$ is around 3\%–15\%, while a degradation appears at 20\%.}
    \label{tab:sensitive1}
\end{table}

\begin{table}[h]
    \centering
    \begin{tabular}{ccccccc}
       \toprule
       $N$ & 5 & 10 & 15 & 20 & 25 & 30 \\
       \midrule
              & 0.39 & 0.65 & 0.78 & 0.86 & 0.83 & 0.87 \\
       \bottomrule
    \end{tabular}
    \caption{\textbf{Sensitivity of N}. We conduct ablation study with different number of Feature Query Tokens. The results show that performance is poor when $N$ is small. When $N$ exceeds 15, performance enters a stable regime. With larger $N$, the performance gains are very limited.}
    \label{tab:sensitive2}
\end{table}

\begin{table}[h]
    \centering
    \begin{tabular}{ccc}
       \toprule
        Fold Shirt & Calculating & Wrapper \\
       \midrule
        0.035 & 0.036 & 0.070 \\
       \bottomrule
    \end{tabular}
    \caption{\textbf{Sensitivity of random seed}. We conduct ablation study with 5 different random seeds. The results show that our method is not sensitive to seed variation.}
    \label{tab:sensitive3}
\end{table}

As can be seen from the results in Table~\ref{tab:sensitive1}, performance remains relatively stable when $\tau$ is around 3\%–15\%, while a degradation appears at 20\%. We find this degradation mainly arises in scenarios requiring complex decision-making, where the model exhibits less stable behavior and occasionally produces inaccurate actions. This is intuitive, since masking too many tokens can impair the completeness of the control intention, thereby leading to reduced performance. Nevertheless, our method provides a relatively wide range of stable performance, indicating that it is not particularly sensitive to the choice of $\tau$.

The results in Table~\ref{tab:sensitive2} show that when $N$ is small, model performance is relatively poor, mainly due to an information bottleneck that makes robust decision-making difficult. However, when $N \ge 15$, performance enters a stable regime. We also note that further increasing $N$ leads to only limited performance gains but larger training cost, and may even result in performance degradation with insufficient data.

Regarding different random seeds, as shown in Table~\ref{tab:sensitive3}, we observe that our method is not sensitive to seed variation.

% We hope our study can offer insights and practical experience for the development and training of other VLA models in the community.

\section{Limitations}
Although our algorithm has been empirically demonstrated to be effective, a rigorous theoretical guarantee has yet to be established. This constitutes a current limitation of our work and remains an important open problem in the VLA literature. We will devote our future efforts to establishing a rigorous theoretical foundation for the proposed method.

\section{Conclusion}
In this paper, we propose \textbf{SyVLA}, a VLA model that achieves a balance between robotic task competence and multi-modal capabilities. In a Think-Before-Act manner, it can follow complex user instructions and accomplish long-horizon tasks. To address the entanglement between implicit control intention representations and high-level reasoning representations introduced by this manner, we propose an \textbf{Intention Decoupling} algorithm to perform disentanglement, and theoretically analyze its effectiveness. To further boost the performance of SyVLA, we also develop a \textbf{Similar Sample Guided RL pipeline}, which enables stable reinforcement learning updates in real-world environment. Experimental results demonstrate that our model achieves strong performance on robotic tasks and exhibits better generalization than baselines. Meanwhile, SyVLA also shows excellent vision–language capability, outperforming other VLA models. We will fully open source the implementation details and training data of SyVLA. To the best of our knowledge, this is among the first fully open-sourced VLA models in the community that possess vision–language understanding and reasoning capabilities. % We hope our work and insights will facilitate the development of subsequent general-purpose embodied intelligence models.

\newpage

\section*{Acknowledgments}
This work is supported by New Generation Artificial Intelligence-National Science and Technology Major Project (No. 2025ZD0122901). This work is also supported by National Science Foundation of China (No. No.62572313, No.62106139).

\section*{Impact Statement}
This paper presents work whose goal is to advance the field of robotics and embodied intelligence. There are many potential societal consequences of our work, none of which we feel must be specifically highlighted here.


\bibliography{main}
\bibliographystyle{icml2026}

%%%%%%%%%%%%%%%%%%%%%%%%%%%%%%%%%%%%%%%%%%%%%%%%%%%%%%%%%%%%%%%%%%%%%%%%%%%%%%%
%%%%%%%%%%%%%%%%%%%%%%%%%%%%%%%%%%%%%%%%%%%%%%%%%%%%%%%%%%%%%%%%%%%%%%%%%%%%%%%
% APPENDIX
%%%%%%%%%%%%%%%%%%%%%%%%%%%%%%%%%%%%%%%%%%%%%%%%%%%%%%%%%%%%%%%%%%%%%%%%%%%%%%%
%%%%%%%%%%%%%%%%%%%%%%%%%%%%%%%%%%%%%%%%%%%%%%%%%%%%%%%%%%%%%%%%%%%%%%%%%%%%%%%
\newpage

\appendix

\onecolumn

\section{More Implementation Details}\label{app:implement}
We describe the implementation details of our SyVLA model.

We choose Qwen2.5-VL 3B as the VLM of SyVLA. For the Feature Query Tokens, we use a fixed set of learnable tensors. Across all three experiments, we use 20 Feature Query Tokens with a dimensionality of 2048, matching the hidden state dimensionality of the VLM. We build our Action Expert based on a transformer architecture, which is a 0.69B parameter Flow Matching model. Concretely, it consists of a 15-layer decoder-only transformer with hidden size 1408, 32 attention heads, and an MLP whose ff\_ratio is 2.67 in each layer. To support the state-of-the-art dual-system architecture and asynchronous inference, we additionally equip the Action Expert with a Siglip ~\cite{zhai2023sigmoid} network to encode visual observations. To enable pretraining on cross embodiment data, where the action and state have different dimensionalities across datasets, we maintain a dataset-specific MLP-based Encoder–Decoder pair for each dataset. Following common practice in VLA, we first normalize the raw input actions and states using dataset mean and variance, then use the dataset-specific Encoder to project them to the Action Expert's hidden size. After the transformer forward pass, we use the Decoder to project the predicted actions embedding back to the original dimensionality, and then unnormalize them using the mean and variance to obtain the actual actions. An example input sequence to the Action Expert transformer includes: low-resolution tri-view image features produced by Siglip, joint position features of the robotic arm embedded by the dataset-specific Encoder, an embedding of the current flow matching decoding timestep, and an embedding of the current noisy action encoded by the action Encoder. We want the Action Expert to closely follow the control intention produced by the VLM; therefore, at every transformer layer, we add a cross-attention module that performs standard cross attention between the transformer input sequence and the control condition. To increase the control frequency of the VLA model, we follow common practice and use action chunk method, i.e., the VLA model predicts multiple steps of actions at once and executes them in a row. Throughout all experiments, we consistently use an action chunk length of 50.

A point worth emphasizing is that the Feature Query States produced by the VLM are not used directly as the control condition $\mathbf{C}$ for the Action Expert; instead, they are first passed through an MLP Adapter. This corresponds to what we describe in Section~\ref{sec:3.1} and~\ref{sec:3.2}, i.e., \( \mathbf{C} = \mathrm{Adapter}(\mathbf{H}) \). We do so for two reasons: (1) to change the dimensionality of the Feature Query States to match the Action Expert hidden size, and (2) to map the Feature Query States into a more suitable space so that the Action Expert can better leverage them for decision making.

Our training procedure has three stages: pretraining, task fine-tuning, and reinforcement learning.

\paragraph{Pretraining} In the pretraining stage, we conduct mixed training on a large-scale robot dataset of roughly 450 hours (if 25 Hz) and several multi-modal datasets. Specifically, our robot data mainly consists of a subset of Open X Embodiment ~\cite{open_x_embodiment_rt_x_2023}, a subset of RH20T ~\cite{fang2024rh20t}, and a portion collected by ourselves, where our self-collected data accounts for less than 5\% of the total robotic data. We clarify that although Open X Embodiment and RH20T are large-scale open-source robot datasets, our total data volume is still far smaller than industrial VLA models such as Pi0 and GR00T, and is less than 5\% of the 10k-hour scale claimed by Pi0. This is mainly because many low-quality or corrupted samples in the open-source datasets were removed during our data cleaning stage. In our robot dataset, only about 1\% is annotated with chain-of-thought(CoT); nevertheless, we find that such limited CoT data, together with mixed training on multi-modal datasets, can still effectively endow the VLA model with basic "Think-Before-Act" capability. For the remaining unannotated robotic data, we disable the "Think-Before-Act" manner during training and let the model directly output Feature Query States $\mathbf{H}$.

Our multi-modal datasets mainly include COCO 2017 ~\cite{lin2014microsoft}, VQA-v2 ~\cite{balanced_vqa_v2}, Pixmo Point Explanations ~\cite{deitke2025molmo}, Pixmo Points ~\cite{deitke2025molmo}, and Cambrain ~\cite{tong2024cambrian1}. To improve multi-modal data quality, we perform fine-grained cleaning on these datasets. During the entire pretraining process, we control the mixing ratio of multi-modal data to 35\% via a Batch Sampler. We train for about 1.5 days using 32 H200 GPUs with Deepspeed Zero2 ~\cite{rajbhandari2020zero}.

\paragraph{Task Fine-tuning} In the task fine-tuning stage, following the standard practice in VLA, we collect several hundred trajectories for each target task and perform detailed CoT annotation. Then, starting from the pretrained SyVLA model, we further train it using the task-specific dataset mixed with the same multi-modal datasets used in pretraining, teaching SyVLA how to solve the target task. After this stage, our VLA model can complete the task with a reasonably good success rate, while still retaining substantial visual-language understanding and reasoning capabilities. In this stage, we train using 8 H100 GPUs with Deepspeed Zero2, and reduce the multi-modal mixing ratio to around 30\%.

Other major hyperparameters for the pretraining and task fine-tuning stages are provided in the Table~\ref{tab:hyper-param}.

\begin{table}[h]
    \centering
    \begin{tabular}{ccc}
       \toprule
       Parameter & Pretraining & Task Fine-tuning \\
       \midrule
       training time & 1.5 days & 6-18h \\
       batch size     & 512     & 32   \\
       \midrule
       learning rate type & cosine with warm up & cosine with warm up \\
       wram up steps & 5000 & 1000 \\
       peak lr  & 1e-4    & 3e-5 \\
       decay lr & 1e-5    & 3e-6 \\
       \midrule
       optimizer & AdamW & AdamW \\
       beta 1 & 0.9 & 0.9 \\
       beta 2 & 0.999 & 0.995 \\
       eps & 1e-8 & 1e-8 \\
       \midrule
       clip gradient norm & 1.0 & 1.0 \\
       \bottomrule
    \end{tabular}
    \caption{\textbf{Major hyper-parameters during pretraining and task fine-tuning}}
    \label{tab:hyper-param}
\end{table}

\begin{figure}[h]
    \centering
    \includegraphics[width=0.8\linewidth]{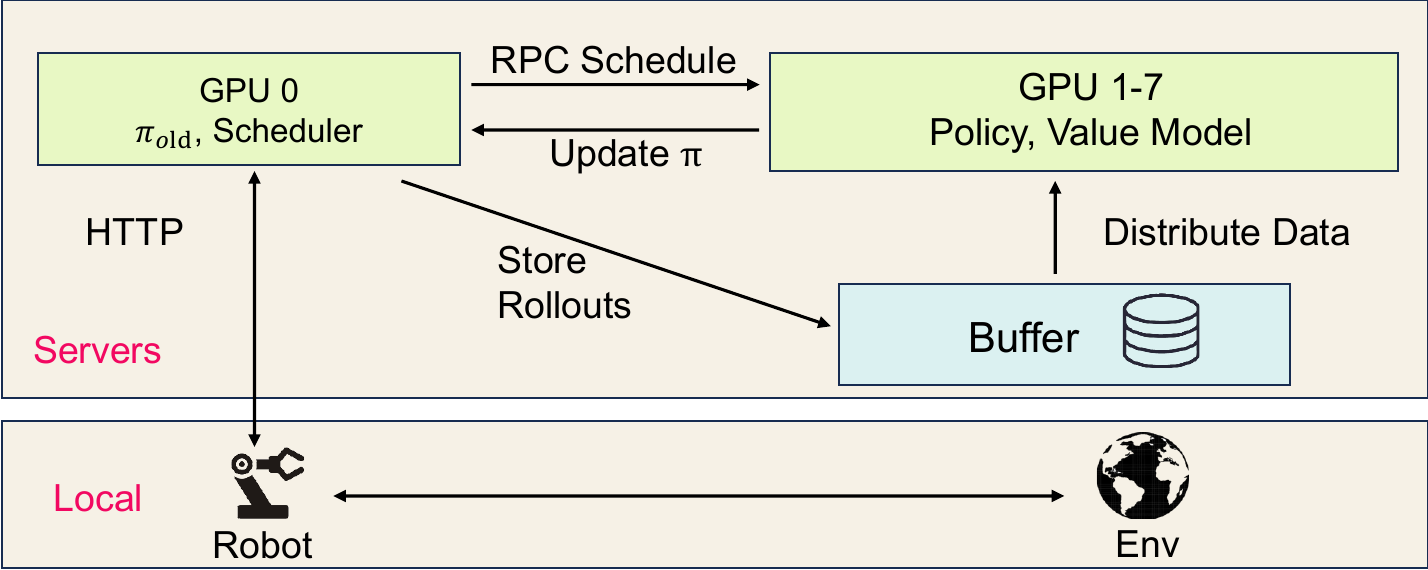}
    \caption{\textbf{The architecture of our RL infrastructure}}
    \label{fig:rl_infra}
\end{figure}

\paragraph{Reinforcement Learning} In the reinforcement learning stage, we perform RL starting from the SyVLA model obtained in the previous stage. We build a value model of about 100M parameters with a Siglip encoder and a transformer network to predict the value of current state based on observation. To mitigate the prediction noise of the value model in early RL, we warm-up it using the imitation-learning(IL) dataset from task fine-tuning stage. Since the IL dataset is fully offline, we directly compute standard state values via Monte Carlo returns to accelerate warm-up training. After obtaining the warmed-up value model, we choose PPO ~\cite{schulman2017proximal} as the base algorithm and adapt it to a large degree to fit our algorithm and start RL training.

To enable efficient and stable RL, we implement an RL Infra as illustrated in Figure~\ref{fig:rl_infra}. Concretely, we use Http communication between the server and the local robot: the robot’s observation images, proprioceptive states, and the user instruction are sent to the server via Http for inference; the server returns the actions to be executed; the robot executes them and repeats this loop. After receiving an inference request, while performing VLA inference, the server also writes observation images into a Replay Buffer (here we use disk as the Replay Buffer), which is used for RL training. On the server side, we split 8 GPUs into two roles: GPU0 acts as the \textbf{Scheduler}, hosting the old policy, providing Http inference service, saving rollout data, and using RPC to schedule the other 7 GPUs; GPUs 1–7 act as \textbf{Workers} responsible for actual RL training. On GPUs 1–7, we load both the value model and the policy model, and perform RL updates using the rollout data saved by the Scheduler together with the Expert dataset from the imitation-learning stage. After the Workers have updated for a certain number of steps, GPU0 receives a synchronization signal and replaces its old policy with the new policy provided by the Workers. Since the stage-two SyVLA model can already complete the task with a non-trivial success rate, we use a 0–1 reward and do not encounter severe reward sparsity. Rewards are provided by humans and are also sent via Http to the Scheduler on the server and recorded into the rollout dataset. We emphasize that this does not incur heavy extra human labor: real-world RL entails certain risks and typically requires human supervision and frequent rollout recovery. While supervising the policy, a human expert can submit whether a rollout succeeds using a few keyboard presses, which is lightweight.

We use the Similar Sample Guided RL algorithm introduced in Section~\ref{sec:3.3} to improve RL stability. Specifically, we use a CLIP model ~\cite{radford2021learning} to encode multi-view observation images into feature vectors for each timestep, and compute sample similarity via weighted cosine similarity across views. Since our robotic arm uses three views, we assign a weight of 0.83 to the center camera view and 0.083 to each of the left and right wrist camera views. We can pre-encode the IL dataset with multiprocessing and store it as a fixed vector retrieval database, so at RL time we only need to compute features for each rollout sample and query the database, which does not introduce substantial additional compute overhead. We merge the two data sources into a single batch for policy optimization. For each task, we allow at most 10 RL episodes, and each episode contains at most 20 rollout trajectories.

A key issue, however, is that we find assigning advantages to Expert data using the standard GAE formula to guide policy-gradient updates leads to rapid gradient explosion (within 1000 update steps). This phenomenon is consistently observed across multiple runs with different random seeds. We further find that even if we completely stop using rollout updates and only perform standard GAE updates on the expert dataset, the same issue occurs. We believe this comes from incorrect advantage assignment, which forces the model to distinguish “good” versus “bad” actions within the expert dataset, confusing the model. Therefore, we abandon the standard PPO GAE for expert data and instead assign a constant advantage of 1.0 to all expert samples, which means we want the policy to follow expert behavior. In addition, we clip the GAE values for rollout data by truncating them to the range \([-0.8, +0.8]\), reflecting that the value estimates have limited confidence and acknowledging that expert data provides stronger learning signals.

Combining the above, we observe that this strategy substantially stabilizes the RL process.
\section{Detailed Theorem Proof}\label{app:theorem}
In this section we offer a detailed proof of our theorems in Section~\ref{sec:3.2}.

We revisit our approach in Sections~\ref{sec:3.1} and Section~\ref{sec:3.2}. Feature Query Tokens are appended after the CoT autoregressively generated by the VLM model and, through a forward pass, produce the Feature Query States $\mathbf{H}_{\mathrm{raw}}$. Then, $\mathbf{H}_{\mathrm{raw}}$ is fed into the Action Expert, where it is first processed by an MLP Adapter to yield the control condition $\mathbf{C}$ for flow matching. Together with other necessary conditions, $\mathbf{C}$ is used by the Action Expert to predict the actions, which are used to compute the action loss $L_{\mathrm{action}}$ with action label. Our Intention Decoupling algorithm computes the gradient of $L_{\mathrm{action}}$ with respect to each Feature Query State, $\mathbf{G}=\{\mathbf{g}^0, \mathbf{g}^1, \cdots, \mathbf{g}^{n-1}\}$, where $\mathbf{g}^i$ denotes the gradient of $L_{\mathrm{action}}$ with respect to the $i$-th Feature Query State $ \mathbf{h}_{\mathrm{raw}}^i$. We select those $\mathbf{g}^i$ whose $\ell_2$-norm is below the $q$-quantile threshold (we use $\tau$ to denote this threshold) of $\mathbf{G}$, and mask the corresponding $i$-th Feature Query State to 0.

For simplification, we use $h_i$ to denote the $i$-th Feature Query State $h_{raw}^i$ below. 

\subsection{Effectiveness of our algorithm}
The attention layer can be denoted as the following.
\begin{theorem}
For each input query vector $q$, the result of attention layer $z$ is
\[
    z = \hat{f}(q) = \frac{\sum_{i=1}^n \kappa(q, k_i)v_i}{\sum_{i=1}^n \kappa(q, k_i)}
\]
, where $\kappa(q, k) = exp(\frac{q k^T}{\sqrt{d_k}})$ is the kernel function, and $d_k$ is the rescale factor.
\end{theorem}

In our Action Expert, there is a fixed action query token at the end of input sequence, which is the query $q$ in the above theorem.

Using the above theorem and assuming the input has already decentralized, we can get that a single-layer transformer Action Expert can be written as:

\[
    \hat{a} = \phi(\mathbf{H}) = \psi(\frac{\sum_{i=1}^n \kappa(q, W^K h_i)W^V h_i}{\sum_{i=1}^n \kappa(q, W^K h_i)})
\]
, where $W^k, W^v$ are the kv mapping matrices of the self-attention layer, $h_i$ is the $i$-th Feature Query State, $\psi$ is a non-linear function. 

We use $\alpha_i = \kappa(q, W^Kh_i)$ to denote the $i$-th kernel function value, and $v_i = W^V h_i$ to denote the $i$-th value vector. Use $z(H)$ to denote the intermediate variable $\frac{\sum_{i=1}^n \alpha_i \cdot v_i}{\sum_{i=1}^n \alpha_i}$, then $\hat{a} = \psi(z(H))$. we have

\begin{align*}
    \frac{\partial \ell}{\partial h_i} &= \frac{\partial \ell}{\partial z} \cdot \frac{\partial z}{\partial h_i} \\
    &= \frac{\partial \ell}{\partial z} \cdot \frac{1}{(\sum_{j=1}^n \alpha_j)^2} \left[ \frac{\partial (\alpha_i v_i)}{\partial h_i} \cdot (\sum_{j=1}^n \alpha_j) - (\sum_{j=1}^n \alpha_j v_j) \cdot (\frac{\partial \alpha_i}{\partial h_i})^T\right] \\
    &= \frac{\partial \ell}{\partial z} \cdot \frac{1}{(\sum_{j=1}^n \alpha_j)^2} \left[ v_i (\frac{\partial \alpha_i}{\partial h_i})^T \cdot (\sum_{j=1}^n \alpha_j) + \frac{\partial v_i}{\partial h_i} \alpha_i \cdot (\sum_{j=1}^n \alpha_j) - (\sum_{j=1}^n \alpha_j v_j) \cdot (\frac{\partial \alpha_i}{\partial h_i})^T \right] \\
    &= \frac{\partial \ell}{\partial z} \cdot \frac{1}{(\sum_{j=1}^n \alpha_j)^2} \left[ \sum_{j=1}^n \alpha_j (v_i - v_j) \cdot (\frac{\partial \alpha_i}{\partial h_i})^T + \frac{\partial v_i}{\partial h_i} \alpha_i (\sum_{j=1}^n \alpha_j) \right]
\end{align*}

Since $\frac{\partial v_i}{\partial h_i} = W^v$, $\frac{\partial \alpha_i}{\partial h_i} = \frac{\partial exp(q(W^k h_i)^T)}{\partial h_i} = \alpha_i (W^k)^T q$, the above equation can be written as
\begin{align*}
    \frac{\partial \ell}{\partial h_i} &= \frac{\partial \ell}{\partial z} \cdot \frac{1}{(\sum_{j=1}^n \alpha_j)^2} \left[ \sum_{j=1}^n \alpha_j (v_i - v_j) \cdot (\frac{\partial \alpha_i}{\partial h_i})^T + \frac{\partial v_i}{\partial h_i} \alpha_i (\sum_{j=1}^n \alpha_j) \right] \\
    &= \frac{\partial \ell}{\partial z} \cdot \frac{1}{(\sum_{j=1}^n \alpha_j)^2} \left[ \alpha_i \cdot \sum_{j=1}^n \alpha_j (v_i - v_j) \cdot (q^T W^k) + W^v \alpha_i (\sum_{j=1}^n \alpha_j) \right] \\
    &= \frac{\partial \ell}{\partial z} \cdot \frac{1}{(\sum_{j=1}^n \alpha_j)^2} \left[ \alpha_i \cdot \sum_{j=1}^n \alpha_j \left((v_i-v_j) \cdot q^T W^k + W^v \right)\right] \\
    &= \frac{\partial \ell}{\partial z} \cdot \frac{1}{(\sum_{j=1}^n \alpha_j)^2} \cdot \alpha_i \cdot \left(\sum_{j=1}^n \alpha_j (v_i-v_j) \cdot q^T W^k + \sum_{j=1}^n \alpha_j W^v \right) \\
    &= \frac{\partial \ell}{\partial z} \cdot \frac{1}{\sum_{j=1}^n \alpha_j} \cdot \alpha_i \cdot \left[ \left( v_i - \sum_{j=1}^n \frac{\alpha_j}{\sum_{k=1}^n\alpha_k} v_j \right) \cdot q^T W^k + W^v \right] \\
    &= \frac{\partial \ell}{\partial z} \cdot \frac{1}{\sum_{j=1}^n \alpha_j} \cdot \alpha_i \cdot \left[ \left( v_i - z \right) \cdot q^T W^k + W^v \right]
\end{align*}

Since the $W^k, W_v, q$ are the same for all input Feature Query States $h_i$, we can get that the value of $\frac{\partial \ell}{\partial h_i}$ is determined by two factors: the magnitude of the kernel-function value \(\alpha_i\), and the distance between \(v_i\) and the output variable of attention layer $z$.

For the first case, since \( \alpha_i \) denotes the attention score assigned to the \(i\)-th input Feature Query State during action generation, a small \( \alpha_i \) indicates that the model considers the \(i\)-th input to be irrelevant to the decision.

For the second case, when \(v_i\) is close to the attention layer’s output \(z\), it indicates that the information in \(v_i\) is largely already covered by other Feature Query Tokens, and thus can be regarded as redundant. However, keeping \(v_i\) for decision-making may cause the model to learn a shortcut decision path and ignore the true causal relevance. This could be acceptable under in-domain cases since it doesn't harm the output, while in OoD cases this could lead to non-causal decision process and degraded performance.

\section{More Experiments and Insights}
In this section, we provide additional experimental results and further insights based on extended analyses.  

\subsection{Simulation Performance}
We conduct experiments on the LIBERO benchmark~\cite{liu2023libero} and report success rates. Specifically, we fine-tune the SyVLA model pretrained in Stage 1 on the LIBERO dataset and subsequently evaluate it in the environment. Following Pi0's protocol, we jointly train on data from all four task suites and then directly test on the four tasks. The results are summarized in Table~\ref{tab:libero}. 

\begin{table}[h]
    \centering
    \begin{tabular}{cccc}
       \toprule
       Spatial & Object & Goal & Long \\
       \midrule
       87.7 & 84.0 & 87.3 & 65.7 \\
       \bottomrule
    \end{tabular}
    \caption{\textbf{Result on Libero benchmark}.}
    \label{tab:libero}
\end{table}

Notably, there remains a non-negligible gap between the fidelity of dynamics in the LIBERO simulator and that of the real world, particularly for the deformable-object and fluid-manipulation tasks highlighted in our main experiments.

\subsection{Ablation on Multi-modal Benchmarks}
To better understand the performance of our method on multimodal benchmarks, using the same model architecture and training recipe, we trained and evaluated the model under four different settings to examine its multimodal capabilities. The results are reported in Table~\ref{tab:ablation_mm}.

\begin{table}[h]
    \centering
    \begin{tabular}{cccccc}
       \toprule
        Setting & DocVQA & AI2D & MMMU & MME & HallBench \\
        \midrule
        All (Ours) & 80.01 & 67.7 & 35.78 & 1795 & 42.53 \\
        w/o Intention Decoupling & 63.62 & 71.76 & 38.67 & 1648.12 & 41.7 \\
        w/o Mixed Data & 0.675 & 0.365 & 0.091 & 1255.34 & 22.25 \\
        w/o Both & 0 & 0.0625 & 0.018 & 1,237.83 & 18.71 \\
       \bottomrule
    \end{tabular}
    \caption{\textbf{Ablation Study on multimodal benchmarks}.}
    \label{tab:ablation_mm}
\end{table}

Without mixed training on multimodal datasets, the model suffers from catastrophic forgetting, yielding near-zero performance on three benchmarks and substantial degradation on the other two. Removing the Intention Decoupling algorithm leads to an overall slight performance drop, but the effect is not significant.

\subsection{Motivation on RL designs}
In this section, we provide deeper insights and discussion regarding the design of our RL module.

Our RL strategy is built on imitation learning because RL is particularly difficult for sparse-reward, long-horizon tasks, especially with billion-parameter VLA models. Imitation learning provides a strong initialization with a non-trivial success rate, while the gap between the IL objective and real-world task success motivates the use of RL.

Specifically, we first pretrain and task-finetune SyVLA to obtain a reasonably capable policy, and then further improve it with RL. We do not use an auxiliary BC loss during RL; instead, we adopt our proposed Similar-Sample Guidance method.

Our comparison with other RL methods is motivated by three considerations.

First, sparse rewards, long horizons, and billion-parameter space make RL highly prone to policy collapse. To address this, our method explicitly incorporates the IL dataset to preserve basic capabilities during RL, which is crucial for stable optimization. In contrast, directly applying standard methods such as PPO or SAC can lead to unstable updates, after which all rollouts fail and RL becomes ineffective.

Second, real-world RL is much more constrained by policy improvement speed and sample efficiency than simulation. Because real-world rollouts are costly and limited, value estimation is less reliable. To improve data efficiency, we incorporate IL data and assign fixed advantages to these samples, reducing optimization errors caused by inaccurate value estimates.

Finally, we observe that the main failure mode of IL-trained VLA models is insufficient fine-grained action precision. These models often generate broadly correct trajectories but still fail due to millimeter-level errors in the final stage, especially for flexible objects. Motivated by this, Similar-Sample Guidance RL aims to teach the model the boundary between acceptable and unacceptable actions, which standard RL often fails to capture. By leveraging advantage differences between similar samples, it helps distinguish, for example, between a 2mm error that remains acceptable and a 3mm error that causes failure. This could be an explanation of the additional gains from similar samples in our ablation study.

\subsection{More RL Results}
We reported ablation studies only on Task 1 in Section~\ref{tab:ablation} because we found that it best reflects subtle changes in model performance under all settings, as it imposes the most substantial demands on dexterous manipulation ability. On the other two tasks, the conclusions are consistent with those reported in the paper as well, but the performance variations were smaller. Specifically, the results are shown in Table~\ref{tab:more_rl_ablation}.

\begin{table}[h]
    \centering
    \begin{tabular}{ccc}
       \toprule
       & Task 2 & Task 3 \\
        \midrule
        w/o Similar Retrival & 0.56 & 0.571 \\
        SyVLA & 0.68 & 0.643 \\
       \bottomrule
    \end{tabular}
    \caption{\textbf{Ablation study of RL algorithm on the other 2 Tasks.}}
    \label{tab:more_rl_ablation}
\end{table}

We also compare our \textbf{Similar-Sample Guidance RL} algorithm with recent SOTA RL method SimpleVLA-RL~\cite{li2025simplevla} and Pi-RL~\cite{chen2025pi_} on \textbf{Fold Shirt} task. The results are shown in Table~\ref{tab:more_rl_comp}

\begin{table}[h]
    \centering
    \begin{tabular}{ccc}
       \toprule
        Pi-RL & Simple VLA & SyVLA \\
        \midrule
        0.714 & 0.357 & 0.86 \\
       \bottomrule
    \end{tabular}
    \caption{\textbf{Comparison between our method with other recent rl baselines.}}
    \label{tab:more_rl_comp}
\end{table}

The results show that SimpleVLA-RL performs poorly and sometimes suffers from gradient explosion, consistent with our observations in paper when directly applying PPO. We attribute this to its removal of the KL-divergence constraint, which makes training prone to collapse. In addition, SimpleVLA-RL does not officially support flow matching VLAs. The lack of algorithmic adaptation may also contribute to its poor performance.

Pi-RL adopts a dual-MDP formulation and is compatible with flow-matching models, but mainly relies on PPO-style clipping for policy stabilization. In our experiments, it frequently caused collisions with the table, posing risks such as motor damage or protective shutdowns. We speculate this may stem from its learnable stochastic noise injection, which can introduce unsafe actions during rollout. Notably, Pi-RL reports only simulation results and lacks direct real-world RL evaluation, which may partly explain its limitations in real-world performance and safety.
\section{Detailed Related Works}
\subsection{VLA Model}
Early VLA models ~\cite{kim2024openvla, zitkovich2023rt, pertsch2025fast} generate discrete action tokens in an auto-regressive manner and decode them into continuous control actions, but this formulation often suffers from limited action precision. In contrast, diffusion policy and flow matching based approaches ~\cite{wen2025diffusionvla, wen2025dexvla, bjorck2025gr00t} synthesize actions through multi-step denoising, enabling finer-grained control and dexterous manipulation; however, they commonly use pretrained VLMs only as initialization, leading to catastrophic forgetting of general vision–language capabilities after VLA training.

Recent works ~\cite{zhouchatvla, zhou2025chatvla, zhai2025igniting, intelligence2025pi_} mix internet-scale vision–language data with robot data to better preserve general knowledge and linguistic competence, allowing models to decompose complex tasks through language and to execute under longer-horizon, abstract instructions. Yet explicit reasoning or subtask decomposition can become entangled with control representations, degrading dexterous performance or causing behavioral stalling. The ChatVLA series ~\cite{zhou2025chatvla, zhouchatvla} introduces additional parameters via an MoE architecture, allocating different competencies to different experts to mitigate this interference, yet limitations remain for dexterous control. Wall-Oss ~\cite{zhai2025igniting} and Pi0.5 ~\cite{intelligence2025pi_} use Fast Token ~\cite{pertsch2025fast} to represent actions at intermediate layers and adopt a two-stage generation paradigm; such methods are typically sensitive to the scale and alignment quality of pretraining data, and may produce meaningless action token generations when data are insufficient.

Our approach leverages Feature Query Token and the Intention Decoupling algorithm to effectively preserve general vision–language capabilities, while mitigating the potential coupling between reasoning processes and control representations.

\subsection{Real-Robot RL}
VLA models are typically trained by regressing expert actions, whereas successful task execution often requires composing hundreds or even thousands of action steps. This mismatch between the training objective and the long-horizon goal in real-world execution leads to error accumulation during closed-loop control: small action errors compound over time, gradually driving the agent’s observations into an out-of-distribution(OoD) regime and ultimately causing task failure. Consequently, reinforcement learning is essential for improving the task-level performance of VLA models ~\cite{intelligence2025pi}.

To mitigate the risks of real-robot exploration, recent work ~\cite{zhu2025wmpo, zhang2025reinforcing, hung2025nora, xiao2025world} has explored using world models as environment simulators, generating post-action observations to replace unsafe on-robot rollouts and enable safer reinforcement learning. However, these approaches are difficult to extend to deformable manipulation (e.g., clothes folding), where generating high-fidelity observations of deformable objects and fluids remains challenging. $\pi^*_{0.6}$ ~\cite{intelligence2025pi} adopts RECAP algorithm ~\cite{frans2025diffusion} for offline reinforcement learning to address rollout safety, but it suffers from slow policy optimization and depends on an exceptionally strong base model. RL-100 ~\cite{lei2025rl} improves capability through a two-stage pipeline—offline reinforcement learning followed by online reinforcement learning—but has only been validated on small diffusion models, and its effectiveness for VLA-scale models remains to be established. Some other methods also attempt to improve the data efficiency and safety through residual policies ~\cite{xiao2025self}, human-in-the-loop correction ~\cite{chen2025conrft, luo2024precise}, morphological symmetry augmentation ~\cite{li2025gr}, carefully designed GRPO variants ~\cite{liu2025flow}, etc. However, these approaches heavily rely on human-designed, task-specific solutions, which substantially limits their transferability. Overall, achieving efficient and stable optimization for VLA models remains an urgent open problem.

To this end, our Similar-Sample Contrast Online RL method substantially improves the stability of online reinforcement learning for VLA models by retrieving similar imitation-learning samples, thereby enabling a practical online RL pipeline for VLA models on real robotic arms.
\section{Detailed Baseline Introduction}\label{app:baseline}
\paragraph{Pi0} Pi0 is a VLA model pretrained on a dataset of over 10k hours. It uses Paligemma ~\cite{beyer2024paligemma} as the VLM backbone and implements a Flow Matching Action Expert with the same number of layers as Paligemma, yielding strong task-execution capability. However, Pi0 does not preserve the inherent capabilities of the VLM, and thus cannot perform Visual QA.

\paragraph{GR00T} The GR00T model is structurally similar to ours, adopting a dual-system design consisting of a VLM and a Diffusion Transformer Action Expert. However, GR00T passes the last hidden states of the entire VLM input sequence to the downstream Action Expert, which incurs substantial computational overhead and is difficult to scale to asynchronous inference. In addition, GR00T does not preserve the inherent capabilities of the VLM and cannot perform CoT task decomposition.

\paragraph{Wall-Oss} Wall-Oss supports CoT-style task decomposition and uses Fast Token as an intermediate representation between the VLM and the Action Expert. During inference, the model first auto-regressively generates Fast Tokens, after which the Action Expert conditions on these Fast Tokens to infer and produce continuous actions. We emphasize that although it preserves vision–language capabilities, it has two disadvantages. First, Fast Token training requires an extremely large amount of data; due to its variable-length nature, insufficient data can lead the model to output meaningless sequences. Second, explicitly decoding the control intention into fast tokens may cause information loss, given the limited size of the Fast Token vocabulary. Moreover, its auto-regressive CoT format is overly lengthy, leading to high inference latency and slow execution on real robotic arms, which deviates from the commonly expected paradigm of general-purpose embodied intelligence.

\paragraph{OpenVLA-oft} OpenVLA-oft does not employ an Action Expert; instead, it trains a final VLA model directly using a single 7B VLM. Specifically, it uses a set of fixed Action Query Tokens, feeds them through the VLM, and decodes the resulting last hidden states into actions via an MLP. Notably, it uses L1 loss for imitation learning. However, VLA models trained with L1 loss often have a lower upper bound in action precision than diffusion-policy or flow-matching-based models.

\paragraph{ChatVLA} ChatVLA creates an MoE replicas for the model’s VLM and initializes the newly created MoE structure with pretrained weights. During inference, it performs hard routing to select the MoE experts according to the input content type(from a task flag input by the user). This design allows the model to better preserve vision–language capabilities, but we find that its dexterous manipulation ability is very limited, despite the increase in parameters.

% \section{More Visualization Results of our tasks}\label{app:vis}
% \begin{figure}[h]
%     \centering
%     \includegraphics[width=1.0\linewidth]{figs/app_task1.pdf}
%     \caption{\textbf{Visualization of task 1}}
%     \label{fig:app_task1}
% \end{figure}

% \begin{figure}[h]
%     \centering
%     \includegraphics[width=1.0\linewidth]{figs/app_task2.pdf}
%     \caption{\textbf{Visualization of task 2}}
%     \label{fig:app_task2}
% \end{figure}

% \begin{figure}[t]
%     \centering
%     \includegraphics[width=1.0\linewidth]{figs/app_task3.pdf}
%     \caption{\textbf{Visualization of task 3}}
%     \label{fig:app_task3}
% \end{figure}

%%%%%%%%%%%%%%%%%%%%%%%%%%%%%%%%%%%%%%%%%%%%%%%%%%%%%%%%%%%%%%%%%%%%%%%%%%%%%%%
%%%%%%%%%%%%%%%%%%%%%%%%%%%%%%%%%%%%%%%%%%%%%%%%%%%%%%%%%%%%%%%%%%%%%%%%%%%%%%%

\end{document}